\documentclass[letterpaper, 10 pt, conference]{ieeeconf}  

\IEEEoverridecommandlockouts                              

\usepackage{graphicx} 
\usepackage{multirow}
\usepackage{hyperref}
\usepackage{listings}
\usepackage{xcolor}
\usepackage{mathabx}

\title{\LARGE \bf
CARMA: Context-Aware Situational Grounding of Human-Robot Group Interactions by Combining Vision-Language Models with Object and Action Recognition
}

\author{Joerg Deigmoeller$^{1}$, 
Stephan Hasler$^{1}$, 
Nakul Agarwal$^{2}$, 
Daniel Tanneberg$^{1}$, 
Anna Belardinelli$^{1}$, \\ 
Reza Ghoddoosian$^{2}$,
Chao Wang$^{1}$, 
Felix Ocker$^{1}$,
Fan Zhang$^{1}$, 
Behzad Dariush$^{2}$, 
Michael Gienger$^{1}$%
\thanks{$^{1}$\textit{Honda Research Institute Europe}, 63073 Offenbach, Germany
        {\tt\small firstname.lastname@honda-ri.de}}%
\thanks{$^{2}$\textit{Honda Research Institute USA}, San Jose, CA 95134, USA
        {\tt\small firstname.lastname@honda-ri.de}}%
}

\begin{document}

\maketitle
\thispagestyle{empty}
\pagestyle{empty}

\begin{abstract}

We introduce CARMA, a system for situational grounding in human-robot group interactions. Effective collaboration in such group settings requires situational awareness based on a consistent representation of present persons and objects coupled with an episodic abstraction of events regarding actors and manipulated objects.
This calls for a clear and consistent assignment of instances, ensuring that robots correctly recognize and track actors, objects, and their interactions over time. To achieve this, CARMA uniquely identifies physical instances of such entities in the real world and organizes them into grounded triplets of actors, objects, and actions. 

To validate our approach, we conducted three experiments, where multiple humans and a robot interact: collaborative pouring, handovers, and sorting. These scenarios allow the assessment of the system's capabilities as to role distinction, multi-actor awareness, and consistent instance identification. Our experiments demonstrate that the system can reliably generate accurate actor-action-object triplets, providing a structured and robust foundation for applications requiring spatiotemporal reasoning and situated decision-making in collaborative settings.

\end{abstract}

\section{INTRODUCTION}
\label{introduction}

Interactive robots are increasingly expected to naturally engage in group interactions involving multiple humans and other robots. Effectively participating in these group settings requires robots to clearly identify and interpret interactions among distinct actors and objects. Yet, precisely how robots can reliably achieve this level of understanding remains an open challenge.

Recent advancements, particularly in Vision Language Models (VLMs), have significantly improved perception in various domains, including robotics, allowing robots to recognize objects and interpret scenes. However, these models typically ground observations at the class level \cite{c22, c23}, which is insufficient for detailed and continuous instance-level understanding required in dynamic multi-agent scenarios.

Proper grounding, in our view, must therefore extend significantly beyond mere entity classification. It involves a tight link between real-world observations and abstract concepts, such as actors or objects. Establishing such link, also called anchoring \cite{c25,c25b}, requires the consideration of two critical aspects \cite{c24}:

\begin{enumerate}
\item appearance-based consistency: reliably recognizing visual similarity, despite visual changes or occlusions.
\item spatial-temporal consistency: ensuring an instance remains uniquely identifiable across time and locations.
\end{enumerate}

Additionally, grounding should not remain limited to individual observations alone. Real-world group interactions inherently involve structured relationships, i.e., interactions among actors, actions, and objects to be captured and represented as coherent triplets (e.g., action patterns \cite{c26}). This kind of \emph{situational grounding} enables robots to reliably identify and re-identify individual entities, and to interpret their dynamic interactions by abstracting observed action events into structured episodes.

With this detailed understanding, we close the loop back to our initial motivation: enabling robots to effectively interpret and meaningfully participate in group interactions. Only by addressing grounding at the level of clearly distinct, disambiguated instances and their relationships, robots can meet the demands of realistic, collaborative environments.

In this paper, we introduce a situational grounding framework, to reduce the gap between continuous visual perception and symbolic, discrete representation to recognize interactions between actors. In this framework, an actor -- which can be either a human or a robot -- is uniquely identified and grounded within interactions with objects and other actors. We propose a novel system architecture by combining: 

\begin{enumerate}
    \item storing instance properties in a memory for instance (re-)identification in the current scene as well as for long-term storage    
    \item representing situational grounded actor-action-object triplets, where actions provide the semantic relationship between objects and actors
    \item disambiguating and uniquely identifying objects and actors in a scene
\end{enumerate}

\section{RELATED WORK}
\label{related_work}


As robotics technology advances, the integration of LLMs and VLMs has become crucial for developing systems that are not only autonomous but also capable of complex interactions with humans. Recent surveys on LLMs and VLMs in robotics \cite{c1, c2} underscore the increasing adoption of multi-modal foundation models to enhance robotic perception, reasoning, and decision-making capabilities. Yet, the challenge remains for robots to seamlessly engage in collaborative human activities, which requires a deep semantic understanding of situational contexts and the precise grounding of physical instances within the environment. 

The approach ``VLM-See-Robot-Do`` \cite{c12} addresses this gap by converting human demonstration videos into robotic actions, utilizing Language Model Programs. This method criticizes the current advancements in VLMs that analyze video sequences \cite{c27, c28, c29}, indicating their insufficient granularity for proper environmental grounding. Instead, it advocates for the analysis of single video frames combined with object detection to enhance accuracy and context relevance. However, this method does not address interactive settings involving multiple actors.

Conversely, in the realm of mobile robotics, models like ``VLM-Social-Nav`` employ VLMs to navigate socially complex environments by identifying key objects and social cues \cite{c7}. Extensions of these models, such as ``VLFM`` and ``ZSON``, combine semantic understanding with goal-directed navigation \cite{c8,c9}. These systems excel in path planning but do not adequately capture purposeful human actions and intentions, which are vital for collaborative scenarios. While they navigate human-centered spaces effectively, their support for direct social interaction and teamwork is limited.

In the field of human-robot interaction, research has been established towards adaptive behavior in group settings. 
The paper \cite{c25b} presents a system that grounds natural language in human-robot interaction using symbolic reasoning and perspective-taking. It enables robots to interpret vague, situated language, however, focusing on dyadic interactions.
This concept of Object-Action Complexes (OACs, \cite{c30}) is a formal framework for grounding and hierarchically organizing sensorimotor experiences into symbolic action representations for autonomous robots, however not including an explicit model of human behavior or interaction.
``LaMI`` \cite{c10} incorporates LLMs for multi-modal interactions, facilitating high-level reasoning in object manipulation and social engagement. Similarly, the Attentive Support framework \cite{c11} introduces proactive assistance by allowing robots to decide when and how to support human groups without disrupting their social interactions. Yet, both approaches rely on marker-based object detection and lack scalability in open-world scenarios. 

Our paper aims to narrow the gap in human-robot group interaction systems by enhancing visual understanding capabilities to accurately encompass scenarios involving multiple actors.

\section{CARMA FRAMEWORK}
\label{frame_work}

\begin{figure*}[thpb]
  \centering
  \includegraphics[width=\linewidth]{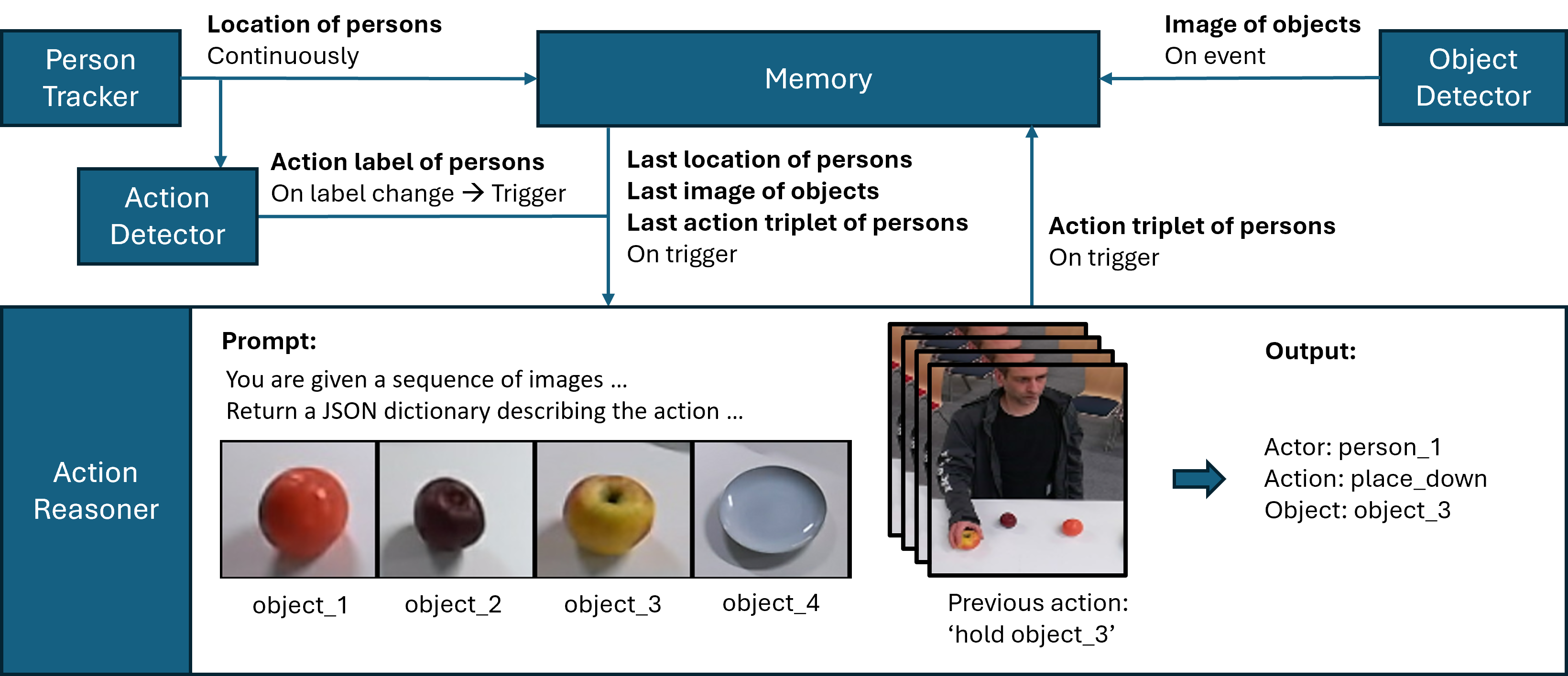}
  \caption{Overview of the proposed framework. The Action Reasoner is a VLM (GPT4o~\cite{c19}) whose prompt has three inputs: (1) A region of the current camera frame based on the location of a person\_y. This serves as the main visual input as it contains the relevant scene entities in their spatial arrangement. (2) An image of each previously detected object\_x. This provides a visual reference to known object instances. (3) The previous action triplet for person\_y. This helps to resolve action label ambiguity by temporal context. -- These inputs are retrieved from the Memory module which acts as an information hub. The Memory is continuously fed with the location of persons by the Person Tracker, while the Object Detector provides a single image for each detected object at system start. A fast low-level Action Detector predicts action labels for each person from the continuous image stream. Whenever a person's action label changes, the Action Reasoner is triggered. The predicted action label can serve as an additional VLM input.}
  \label{figure:carma}
\end{figure*}

In this section, we present the CARMA framework by first focusing on its output: uniquely identified actor--action--object triplets. Each triplet represents who is doing what with which object at any given moment, forming a structure for situational grounding in human-robot group interactions. By organizing real-world events into such triplets, the framework keeps track of distinct actors and objects in a way that remains consistent over time, even as people move around or objects get picked and placed.
To illustrate the functionality of the CARMA framework, consider a collaborative cocktail preparation scenario involving two human actors and one robot. One person pours juice from a bottle into a shaker, while simultaneously the other person slices lemons. Within CARMA, each entity—actors (humans, robot) and objects (bottle, shaker, knife, lemons, ice container)—is continuously identified and grounded uniquely. The resulting interactions are represented as structured actor-action-object triplets, for instance: {actor: human 1, action: pour, object: bottle}, {actor: human 2, action: slice, object: lemon}, and {actor: robot, action: hand over, object: ice container, receiver: human 1}. Compared to traditional class-level grounding, this instance-level situational grounding allows to resolve item ownership among actors, detecting missing ingredients during task execution, and resolving ambiguities arising from multiple similar objects or parallel activities.

In the remainder, we describe how physical instances are detected and grouped into triplets using the five components Memory, Object Detector, Person Tracker, Action Detector and Action Reasoner. As shown in Figure \ref{figure:carma}, the Object Detector, Person Tracker and Action Detector are used to identify individual scene elements, without further context. Those information are fed into memory for consistent storage and retrieval of uniquely identified instances at any stage in the process. Finally, the Action Reasoner combines all individual information on a higher level, considering the situational condition present in the complete image.

\textbf{Memory}. The memory system builds a central element, as it stores all detected object instances, using a MongoDB database. Observed instances are treated as measurements with attached properties such as unique identifiers, cropped instance images, instance locations and time stamps, allowing a re-identification of instances at any later stage in the process.

\textbf{Object Detector.} Objects, stored in memory, are not continuously tracked, rather detected and updated in an event-driven way, for example at system start-up or when interactions have been detected. First, a 3D point cloud clustering is applied to segment the workspace into object candidates. Second, each segmented object is assigned a unique ID and stored in the Memory along with its cropped image and location in 3D. 

\textbf{Person Tracker.} For person actors, consistency over time is ensured through the use of a body pose tracker. The Person Tracker extends the body tracker by assigning a unique ID to each detected individual and forwards cropped person images to the Action Detector. The cropping area in the image is estimated using the 3D body pose projected on the image plane.

\textbf{Action Detector}. To determine when an actor interacts with an object, the Action Reasoner incorporates the Action Detection module that continuously analyzes the visual input provided by the Person Tracker. Specifically, cropped person images are grouped into four-frame sequences per individual and processed through the initial layers of a spatio-temporal model, based on the I3D backbone~\cite{c17}. It returns action labels for each actor independently. The labels serve two complementary purposes. First, they can be used as prior action information, included in the VLM prompt. Second, they can act as trigger signals, any time an action is performed, for more refined interpretation through the Action Reasoner.

\textbf{Action Reasoner}. The Action Reasoner, shown in Figure~\ref{figure:carma} bottom, is the key component providing the final output by combining all outputs of the previously mentioned components into  grounded actor--action--object triplets. The component draws on a VLM, but it is guided by the set of identified actor and object instances in the scene to ensure it remains focused on the most relevant elements. The Action Reasoner uses cropped actor images and object/action data from the Person Tracker, Object Detector, and Action Detector to resolve actor--action--object triplets based on spatial arrangements. In other words, we take advantage of the VLM’s broad language–vision capabilities but ensure the output is grounded in the real-world actors and objects that the system has previously identified.

\begin{figure*}[thpb]
  \centering
  \includegraphics[width=1.05\linewidth]{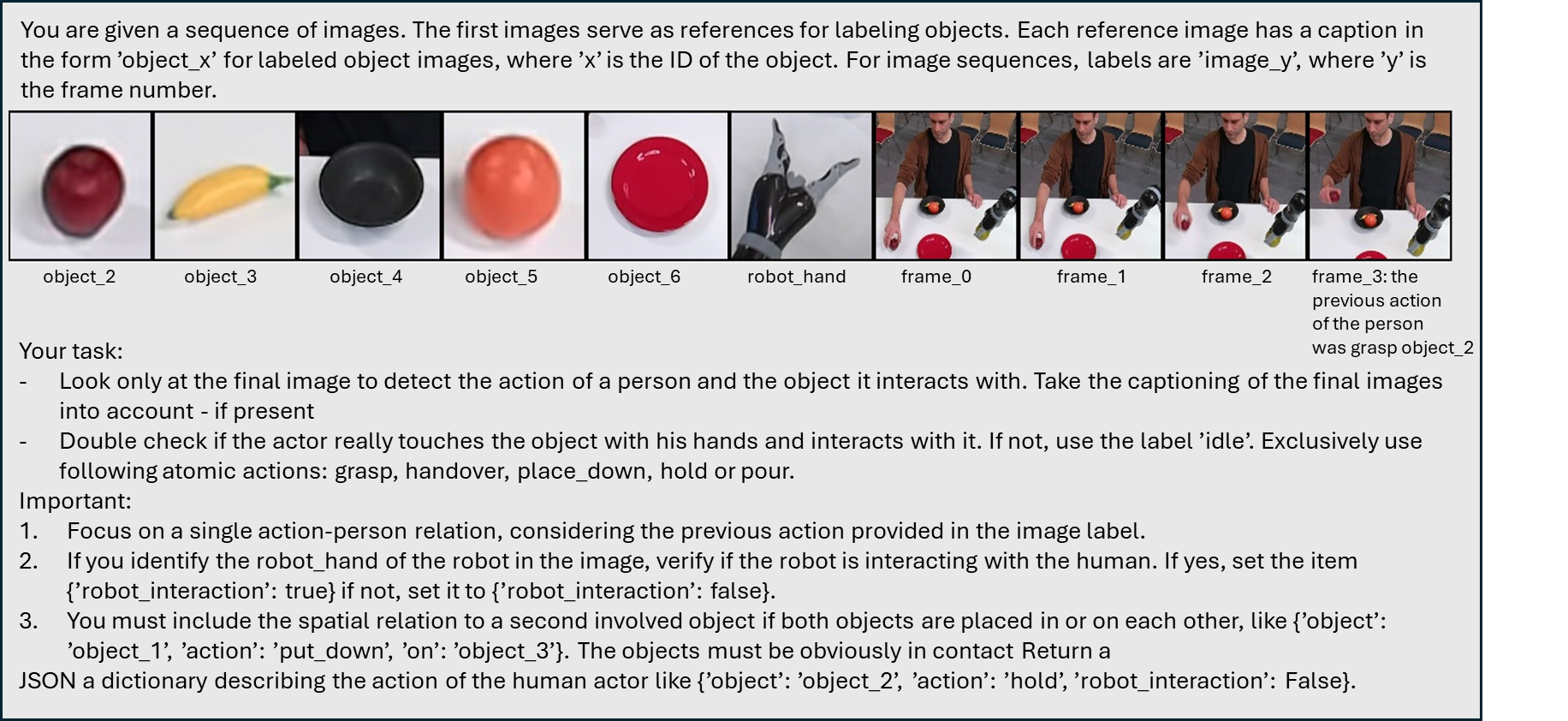}
  \caption{Visualization of the prompt provided to VLM. The prompt begins with a general introduction, followed by cropped object images, each uniquely identifiable via caption. The robot hand is optionally included to assess interaction. The last four images show recent cropped person views before the action trigger; the final image contains the action-object combination from the previous step. The prompt concludes with a task description guiding the VLM through action inference, object assignment, spatial relation, and robot interaction.}
  \label{figure:prompt}
\end{figure*}

Figure~\ref{figure:prompt} shows an example prompt fed into the VLM.
The prompt begins with a general introduction, followed by cropped images of relevant object instances in the scene. Each object is uniquely identifiable to the VLM through its image caption. The robot hand is optionally included as an object to help the VLM determine whether the robot is interacting with the person.
The last four images show the most recent cropped person images that appeared before an action trigger signal. The final image is the one on which the VLM should base its reasoning; it also includes the action-object combination from the previous analysis for this specific actor.
Finally, the VLM prompt includes a detailed task description, which instructs it to: first, infer which action the person is performing; second, assign one of the available objects to the person’s action; third, if applicable, identify a second object involved in the action and describe the spatial relation; and fourth, determine whether the robot hand is interacting with the human.
The final output returned by the Action Reasoner is then of the form:
\{'object': 'object\_2', 'action': 'place\_down', 'on': 'object\_4', 'robot\_interaction': false\}.

For our experiments described below, we used the modules in four different configurations, to investigate their influence on the Action Reasoner:

\begin{enumerate}
\label{enum:conf}
\item \textbf{discrete sampling + previous triplet}: images are sampled at regular time intervals (action label and trigger information is not delivered by the Action Detector, rather the VLM needs to determine actions on its own). The VLM also receives the actor-action-object triplet from the previous time step.
\item \textbf{action trigger + previous triplet}: uses the Action Detector (described in more detail below) to forward only a trigger signal, indicating that a new action has started, without any information about what kind of action it is, i.e. no label. Again, the VLM also receives the actor-action-object triplet from the previous time step.
\item \textbf{action trigger}: Same as 2) this time, the VLM does not receive previous actor-action-object triplet information.
\item \textbf{action trigger + action label + previous triplet}: forward all previously described information plus the action label delivered by the Action Detector.
\end{enumerate}

\section{Evaluation} \label{sec:evaluation}

For the evaluation, we created our own dataset, which is available at \href{https://github.com/HRI-EU/carma}{https://github.com/HRI-EU/carma}. This decision was motivated by two main considerations. Firstly, although existing benchmarks like Video-MME \cite{c18} facilitate large-scale evaluations of VLMs, they primarily assess general context understanding through multiple-choice tests and do not provide the granularity needed for detailed action reasoning within human-robot collaboration scenarios. Second, to the best of our knowledge, no existing dataset provides detailed situational annotations involving multiple interacting actors. The most closely related work \cite{c5} is not publicly available and does not include multi-actor scenarios. Our dataset builds on their experimental setup but extends it to include multi-person interactions and robot involvement. 
Furthermore, we build on their findings, demonstrating that video language models, such as those described in \cite{c27}, perform significantly worse compared to guided VLMs, which provide more structured and context-specific analysis.

\subsection{Experimental Setup}

To validate CARMA’s situational grounding approach, we conducted three interactive table-top experiments:

\begin{enumerate}
\item \textbf{Sorting Fruits}: A banana, two apples, and an orange are sorted into a bowl or onto a plate. 
\item \textbf{Pouring}: A bottle is used to pour liquid into one of several cups. 
\item \textbf{Handover}: Participants hand various items to each other around the table. 
\end{enumerate}
Each task was performed under up to four different settings: 
\begin{itemize} 
\item \textbf{1P}: 1 person performs actions alone. 
\item \textbf{2P}: 2 persons perform actions or interact. 
\item \textbf{1P+R}: 1 person and the robot perform actions independent or interact. 
\item \textbf{2P+R}: Two persons and the robot perform actions independent or interact. 
\end{itemize}

In total, we tested 8 distinct settings, repeated across three runs each, resulting in 24 recordings.

To measure performance, we use the Task Success Rate (TSR) -- the percentage of correctly identified and ordered triplets in comparison to ground truth triplets. TSR considers correct temporal ordering, accurate instance references, correct actor-role attribution, and identification of human–robot interaction events.

\begin{table*}[t]
\caption{Task Success Rate (TSR) based on comparison with ground truth sequence.}
\label{table:tsr}
\begin{center}
\begin{tabular}{|c||c|c|c|c||c|c||c|c||c||}
\hline
 & \multicolumn{4}{c||}{Sorting Fruits} & \multicolumn{2}{c||}{Pouring} & \multicolumn{2}{c||}{Handover} & \multicolumn{1}{c||}{All}\\
& 1P & 2P & 1P+R & 2P+R & 2P & 1P-R & 2P & 1P-R & $\diameter$ \\
\hline
1) discrete sampling + previous triplet & 0.53 & 0.54  & 0.66 & 0.46 & \textbf{0.69} & \textbf{0.8} & 0.65 & 0.55 & 0.61 \\
\hline
2) action trigger + previous triplet & 0.56 & 0.60 & 0.76 & \textbf{0.70} & 0.54 & 0.75 & \textbf{0.71} & \textbf{0.75} & 0.67 \\
\hline
3) action trigger & \textbf{0.69} & \textbf{0.61} & \textbf{0.85} & 0.64 & 0.65 & 0.81 & 0.55 & \textbf{0.75} & \textbf{0.72} \\
\hline
4) action trigger + action label + previous triplet & 0.45 & 0.42 & 0.50 & 0.52 & 0.63 & 0.74 & 0.53 & 0.75 & 0.57 \\
\hline
\end{tabular}
\end{center}
\end{table*}

\begin{table*}[t]
\caption{Accuracy of each role contributing in an action pattern for each experiment. The numbers are averaged over all settings, including one and two persons, as well as the robot. Actors are excluded, as they are always correctly assigned.}
\label{table:roles}
\begin{center}
\begin{tabular}{|c|c|c||c|c||c|c||}
\hline
 & \multicolumn{2}{c||}{Sorting Fruits} & \multicolumn{2}{c||}{Pouring} & \multicolumn{2}{c||}{Handover} \\
\hline
& actions & objects & actions & objects  & actions & objects \\
\hline
1) discrete sampling + previous triplet  & 0.56 & 0.78 & \textbf{0.84} & \textbf{0.97}  & 0.74 & 0.77 \\
\hline
2) action trigger + previous triplet  & 0.66 & 0.86  & 0.77 & 0.91 & \textbf{0.86} & \textbf{0.8} \\
\hline
3) action trigger & \textbf{0.67} & \textbf{0.86}  & 0.79 & 0.98  & 0.78 & 0.81 \\
\hline
4) action trigger + action label + previous triplet  & 0.47 & 0.86  & 0.69 & 0.96  & 0.77 & 0.72 \\
\hline
\end{tabular}
\end{center}
\end{table*}

\subsection{Results}

Table~\ref{table:tsr} presents the TSR results for the four system configurations as outlined end of Section~\ref{frame_work}.

Our analysis reveals that configurations 2 and 3 outperform the other configurations by approximately 10\%. This enhanced performance is primarily to be attributed to the trigger signal from the Action Detector, which effectively reduces false detections by pre-filtering events in the image sequence, rather than performing continuous analysis in 4-image chunks. One exception is the pouring scenario, where the continuous analysis performs slightly better by about 3\%. Table~\ref{table:roles} shows that the actions mainly contribute to the performance drop, indicating that certain actions in a sequence may be missed by the Action Detector.

Further comparison between configurations 2 and 3, which differ in their handling of the actor-action-object triplet—carrying it forward from one prompt to the next—shows some drawbacks. The data suggests that the inclusion of past observations sometimes misleads the VLM decision.

Configuration 4 exhibits the poorest performance, as indicated in Table~\ref{table:roles} where the action role scores are consistently lower. This configuration suffers from the misleading influences of erroneous action labels detected by the Action Detector, suggesting that the VLM may overly rely on this input.

Further errors arise from object occlusions or wrong assignment of object instances, as well as wrongly inferred actions for each person. Even in more complex scenarios, such as Sorting Fruits involving two people and a robot (2P+R), the system maintains performance comparable to less complex setups, suggesting robust object and action assignment when the image area is appropriately cropped to focus on relevant regions. 


An additional mode, not included in our experiment, involves pre-filtering object instances from memory that are visually similar to objects in the currently cropped image surrounding the person, based on cosine distances between image embeddings. Initial results showed no substantial improvements over leaving the filtering to the VLM. However, this method has potential benefits, such as providing precise object positions within the image, where VLMs currently lack. This mode, as well as the full CARMA framework can be explored on our GitHub page at \href{https://github.com/HRI-EU/carma}{https://github.com/HRI-EU/carma}.

\subsection{Discussion}

Our evaluation highlights CARMA’s capability to move beyond class-level recognition and toward grounded instance-level identification—a critical requirement for natural and effective human–robot interaction in shared environments. By disambiguating between specific object instances and attributing actions to individual actors, CARMA enables the robot to form a detailed situational model, which is essential for responsive and context-aware collaboration.

This instance grounding becomes particularly important in multi-party scenarios, where overlapping roles, object re-use, and temporal dependencies complicate scene understanding. The ability to distinguish not just “a cup” but “the cup handed by person A to person B” allows robots to infer intentions and track scene changes to react appropriately. 

Overall, the findings demonstrate that structured, grounded perception is not only feasible with current VLM-based systems but also provides concrete functional benefits for collaborative autonomy.

\section{Conclusion}

CARMA contributes a step toward more intuitive and context-aware human–robot collaboration by introducing a structured, event-driven approach to instance-level scene understanding. Rather than treating perception as a static recognition task, our framework integrates memory, guidance, and selective attention to support meaningful interaction in dynamic, multi-agent environments.

By grounding observed actions in specific object instances and actor roles, CARMA provides robots with the foundations for situational awareness, intention inference, and memory-based interaction strategies. These capabilities are essential for real-world deployment in shared human environments, where ambiguity, variability, and social coordination occur.

Furthermore, instead of overwhelming the system with continuous scene analysis, CARMA attends only to salient events, reflecting a cognitively inspired approach to perception. This reduces latencies, which is vital in time-sensitive interactions like handovers or safety-critical tasks.

Looking ahead, CARMA lays the groundwork for interactive systems that not only see, but also remember, adapt, and respond appropriately, which are key qualities for  collaborative robotic agents.


\end{document}